\documentclass[10pt,twocolumn,letterpaper]{article}

\usepackage{iccv}
\usepackage{times}
\usepackage{epsfig}
\usepackage{graphicx}
\usepackage{amsmath,amssymb,wasysym}

\usepackage{booktabs}
\usepackage{colortbl}
\usepackage{threeparttable}
\usepackage{multirow}

\usepackage{xspace}
\usepackage{xcolor}
\usepackage{mathtools}
\usepackage{bm}
\usepackage{stackengine}
\stackMath

\setstackgap{L}{9pt}

\makeatletter
\newcommand\ensuresingleperiod{\@ifnextchar.{}{.\@\xspace}}
\makeatother

\renewcommand*{\ie}{\emph{i.e.}\@\xspace}
\renewcommand*{\vs}{\emph{vs.}\@\xspace}
\renewcommand*{\eg}{\emph{e.g.}\@\xspace}
\renewcommand*{\etc}{\emph{etc}\ensuresingleperiod}
\newcommand*{\aposteriori}{\emph{a posteriori}\xspace}

\DeclarePairedDelimiterX\Set[1]\{\}{%
  
  #1
}

\newcommand*\makeSet[1]{\mathcal{#1}}

\DeclarePairedDelimiterXPP\Prop[1]{\mathbb{P}}(){}{
   
   #1}

\newcommand\funop[1]{\mathop{{}#1}}

\DeclarePairedDelimiterX{\inner}[2]{\langle}{\rangle}{
  \ifblank{#1}{\:\cdot\:}{#1},\ifblank{#2}{\:\cdot\:}{#2}}

\renewcommand{\vec}[1]{\bm{#1}}

\renewcommand\matrix[1]{\bm{#1}}

\DeclarePairedDelimiterXPP\Lone[1]{}{\lVert}{\rVert}{_1}{\ifblank{#1}{\:\cdot\:}{#1}}
\DeclarePairedDelimiterXPP\Ltwo[1]{}{\lVert}{\rVert}{_2}{\ifblank{#1}{\:\cdot\:}{#1}}
\DeclarePairedDelimiterXPP\Lonetwo[2]{}{\lVert}{\rVert}{_{1,2\ifblank{#2}{}{,#2}}}{\ifblank{#1}{\:\cdot\:}{#1}}
\DeclarePairedDelimiterXPP\Linf[1]{}{\lVert}{\rVert}{_\infty}{\ifblank{#1}{\:\cdot\:}{#1}}
\DeclarePairedDelimiter{\abs}{|}{|}

\newcommand*{\defeq}{\mathrel{\coloneqq}}

\DeclareMathOperator*{\arginf}{arg\,inf}

\DeclareMathOperator*{\argmax}{arg\,max}

\newcommand*\energy{\mathcal{E}}

\newcommand*{\image}{\matrix{I}}

\newcommand*{\class}{c}
\newcommand*{\classSet}{\makeSet{C}}

\newcommand*{\classFreq}{\matrix{P}}

\newcommand*{\pixel}{i}
\newcommand*{\pixelOther}{j}
\newcommand*{\pixelSet}{\makeSet{I}}

\newcommand*{\pc}{{\pixel\class}}

\newcommand*\mask{\matrix{M}}

\newcommand*{\simplex}{\Delta}

\newcommand*\Ed{\energy_\mathrm{d}}
\newcommand*\Edf{\funop{\Ed}}

\newcommand*\Er{\energy_\mathrm{r}}
\newcommand*\Erf{\funop{\Er}}

\newcommand*\ErTV{\energy_\mathrm{r}^{\mathrm{TV}}}

\newcommand*\ErCRF{\energy_\mathrm{r}^{\mathrm{ CRF}}}
\newcommand*\ErCRFf{\funop{\ErCRF}}

\newcommand*\regp{\lambda}
\DeclareMathOperator{\TV}{TV}
\newcommand*{\TVg}{\TV_{\!g}}
\newcommand*{\TVgf}{\funop{\TVg}}

\newcommand*{\appearanceKernel}{\matrix{A}}
\newcommand*{\smoothnessKernel}{\matrix{S}}

\newcommand*{\grad}{\nabla}

\usepackage[acronym]{glossaries}
\newacronym{iou}{IoU}{Intersection over Union}
\newacronym{miou}{mIoU}{Mean Intersection over Union}

\newacronym{wa}{WA}{Weak Annotation}
\newacronym{fa}{FA}{Full Annotation}
\newacronym{hfa}{HFA}{Human Full Annotation}
\newacronym{pfa}{PFA}{Predicted Full Annotation}
\newacronym{pam}{PAM}{Predictive Annotator Model}

\newacronym{dcnn}{DCNN}{Deep Convolutional Neural Network}
\newacronym{wa-dcnn}{WA-DCNN}{Weak Annotation \acrshort{dcnn}}
\newacronym{hfa-dcnn}{HFA-DCNN}{Human Full Annotation \acrshort{dcnn}}
\newacronym{pfa-dcnn}{PFA-DCNN}{Predicted Full Annotation \acrshort{dcnn}}

\newacronym{rf}{RF}{Random Forest}
\newacronym{gmm}{GMM}{Gaussian Mixture Model}
\newacronym{vm}{VM}{Variational Model}
\newacronym{tv}{TV}{Total Variation}
\newacronym{fc-crf}{FC-CRF}{Fully Connected Conditional Random Field}
\newacronym{crf}{CRF}{Conditional Random Field}

\newacronym{ms-coco}{MS-COCO}{Microsoft Common Object in Context}

\newacronym{map}{MAP}{maximum \aposteriori}
\newacronym{tsne}{T-SNE}{t-Distributed Neighbor Embedding}
\newacronym{aspp}{ASPP}{Atrous Spatial Pyramid Pooling}

\newacronym{fft}{FFT}{Fast Fourier Transform}
\newacronym{dct}{DCT}{Discrete Cosine Transform}
\newacronym{asb}{ASB}{Alternating Split Bregman}
\newacronym{admm}{ADMM}{Alternating Direction Method of Multipliers}

\definecolor{global}{RGB}{217,95,2}
\definecolor{local}{RGB}{27,158,119}
\definecolor{combined}{RGB}{117,112,179}
\newcommand{\soft}[1]{$\begingroup\color{#1}\CIRCLE\endgroup$}
\newcommand{\potts}[1]{$\begingroup\color{#1}\blacksquare\endgroup$}
\newcommand{\fccrf}[1]{$\begingroup\color{#1}\blacktriangle\endgroup$}
\newcommand{\cmmnt}[1]{\ignorespaces}

\usepackage[pagebackref=true,breaklinks=true,letterpaper=true,colorlinks,bookmarks=false]{hyperref}

\iccvfinalcopy % *** Uncomment this line for the final submission

\def\iccvPaperID{5032} % *** Enter the ICCV Paper ID here

\ificcvfinal\fi
\begin{document}

%%%%%%%%% TITLE
\title{Towards Closing the Gap in Weakly Supervised Semantic Segmentation with
  DCNNs: Combining Local and Global Models}

\author{Christoph Mayer\\
ETH Z\"urich, Switzerland\\
{\tt\small chmayer@vision.ee.ethz.ch}
\and
Radu Timofte\\
ETH Z\"urich, Switzerland\\
{\tt\small timofer@vision.ee.ethz.ch}
\and
Gr\'egory Paul\\
ETH Z\"urich, Switzerland\\
{\tt\small grpaul@vision.ee.ethz.ch}
}

\maketitle
%\thispagestyle{empty}

%%%%%%%%% ABSTRACT
\begin{abstract}
  Generating training sets for \acrshortpl{dcnn} is a bottleneck for modern
  real-world applications.
  This is a demanding task for applications where
  annotating training data is costly, such as in semantic segmentation.
  In the literature, there is still a gap between the performance achieved by a
  network trained on full and on weak annotations.
  In this paper, we establish a strategy to measure this gap and to identify the
  ingredients necessary to reduce it.

  On scribbles, we establish new state-of-the-art results: we obtain a
  \acrshort{miou} of 75.6\% without, and 75.7\% with \acrshort{crf}
  post-processing.
  We reduce the gap by 64.2\% whereas the current state-of-the-art reduces it
  only by 57.5\%.
  Thanks to a systematic study of the different ingredients involved in the
  weakly supervised scenario and an original experimental strategy, we unravel a
  counter-intuitive mechanism that is simple and amenable to generalisations to
  other weakly-supervised scenarios: averaging poor local predicted annotations
  with the baseline ones and reuse them for training a \acrshort{dcnn} yields
  new state-of-the-art results.

\end{abstract}

%%% Local Variables:
%%% mode: latex
%%% TeX-master: "egpaper"
%%% End:

%  LocalWords:  convolutional variational DeepLab

%%%%%%%%% BODY TEXT
% \input{fig-pipeline}
\section{Introduction}

Semantic segmentation aims at extracting \emph{semantically meaningful} segments
and classify each part into one of the classes predefined by the user. It is a
central problem to computer vision, because it bridges a lower-level task (image
segmentation) to a higher-level one (scene understanding). State-of-the-art
models are data-driven and require for training examples of images together with
the segmentation of the intended classes. Recently, \glspl{dcnn} have achieved
the best performance to date on the public data sets used for comparing
different frameworks in a normalized fashion~\cite{Garcia-Garcia2017}, such as
PASCAL VOC~\cite{Everingham2010} or \acrshort{ms-coco}~\cite{lin2014}.

However, \glspl{dcnn} are greedy in the amount of training data. For semantic
segmentation, providing a training set is a demanding task, because it requires
assigning carefully a label to each pixel in the training set. This poses two
problems for real-world applications of semantic segmentation:
\emph{versatility} and \emph{scalability}. Versatility is an issue when the
classes of interest differ from the ones in the training set: this requires
re-annotating the training images. Scalability is an issue when the number of
training images grows significantly, \ie at the scale of data sets that are
nowadays available and required in real-world applications. A solution to these
issues is to rely on \emph{weak supervision}.

\subsection{Weakly-supervised semantic segmentation}
\label{sec:weak-supervision}

In semantic segmentation, a \emph{full annotation} holds information about the
location, the shape, the spatial relationships between segments, the
co-occurrence of classes, the class of each segment, \etc.
In contrast, \emph{weak annotations} do not provide direct examples of semantic
segments, but offer only partial cues:
image-level tags provide class information~\cite{Pathak2015,Huang2018};
point supervision provides class and approximate location~\cite{Bearman2016},
bounding boxes~\cite{Khoreva2017} provide class, approximate location and
extent;
scribbles~\cite{Lin2016,Vernaza2017,Tang2018} provide class, approximate object
location and extent.
In addition to \emph{things} (\ie with a distinct size and shape, \eg cars,
people), scribbles can also annotate \emph{stuff} (\ie with no specific spatial
extent or shape, \eg road, sky), see~\cite{Forsyth1996}.
On one hand, \glspl{wa} are easier to collect (see~\cite{Bearman2016} for
timings), more versatile, and better for upscaling the training set.
On the other hand, \glspl{wa} are not exhaustive and subject to human annotation
errors.

Therefore, weak supervision requires specific training strategies.
Different strategies exist in the literature.
Hong \emph{et al.}~\cite{Hong2015a} adapt the \gls{dcnn} architecture.
Designing new loss functions to promote weak annotations to full ones is
popular: \cite{Bearman2016} use objectness for point supervision and other
higher-level priors, and \cite{Tang2018} designed a new loss function inspired
by "shallow" segmentation.
Post-processing weak annotations to full ones as an intermediate step to train
an existing \gls{dcnn} is also popular:
bounding boxes propagated by GrabCut~\cite{Papandreou} or other
strategies~\cite{Khoreva2017}, scribbles extended by super-pixels and a
variational model~\cite{Lin2016}.

\subsection{(Semi-)interactive segmentation}
\label{sec:semi-inter-segm}
(Semi-)interactive segmentation is a boundary case of the weakly-supervised
setting where the training set is reduced to a single image.
State-of-the-art interactive segmentation frameworks are in essence
\emph{Bayesian} and \emph{variational}.
In the interactive setting, the amount of training data is low (reduced to the
user inputs, such as bounding boxes~\cite{Rother2004a},
scribbles~\cite{Unger2008,Santner2009}, \etc) and makes the prediction for the
unlabelled pixels uncertain.
Therefore, Bayesian models are an attractive paradigm for this task, as they
allow incorporating prior knowledge (\eg boundary length~\cite{Nieuwenhuis2013},
spatial semantic relation~\cite{Diebold2016a} and
co-occurrence~\cite{ladicky2010}, appearance and smoothness in color
space~\cite{Krahenbuhl2011}) that facilitates the predictions when the semantic
model alone is uncertain.

\subsection{Goals and Contributions}

In this work we tackle the problem of training a \gls{dcnn} for the semantic
segmentation problem in a weakly supervised setting. See~\cite{Hong2017a} for a
recent review and the references therein. Our main focus is on scribble
annotations for which annotation data are available for the PASCAL VOC data
set~\cite{Lin2016}. Among the possible strategies described in
Sec.~\ref{sec:weak-supervision}, we follow the strategy consisting in
post-processing the weak annotations to generate full annotations to train a
subsequent \gls{dcnn} from scribbles. 

Our goal is to identify simple ingredients to reduce the gap between the
baseline accuracy achievable by training the network on the weak annotations
only (lower bound) and the accuracy obtained by training on the fully annotated
training set (upper bound).

Our comprehensive experimental design (Tab.~\ref{table:summary}) allowed us to
identify an unexpected interaction between local and global \glspl{pam} that
conspire to boost the overall accuracy (Fig.~\ref{fig:overview}) and establish
new state-of-the-art results (Tab.~\ref{table:comparison}).

%%% Local Variables:
%%% mode: latex
%%% TeX-master: "egpaper"
%%% End:

%  LocalWords:  convolutional DCNNs scalability DCNN DecoupledNet objectness
%  LocalWords:  GrabCut variational tradeoff ResNet VGG DeepLab atrous

\section{Methods}
\label{sec:methods}

\begin{figure}
  \centering
  \includegraphics[width=\columnwidth]{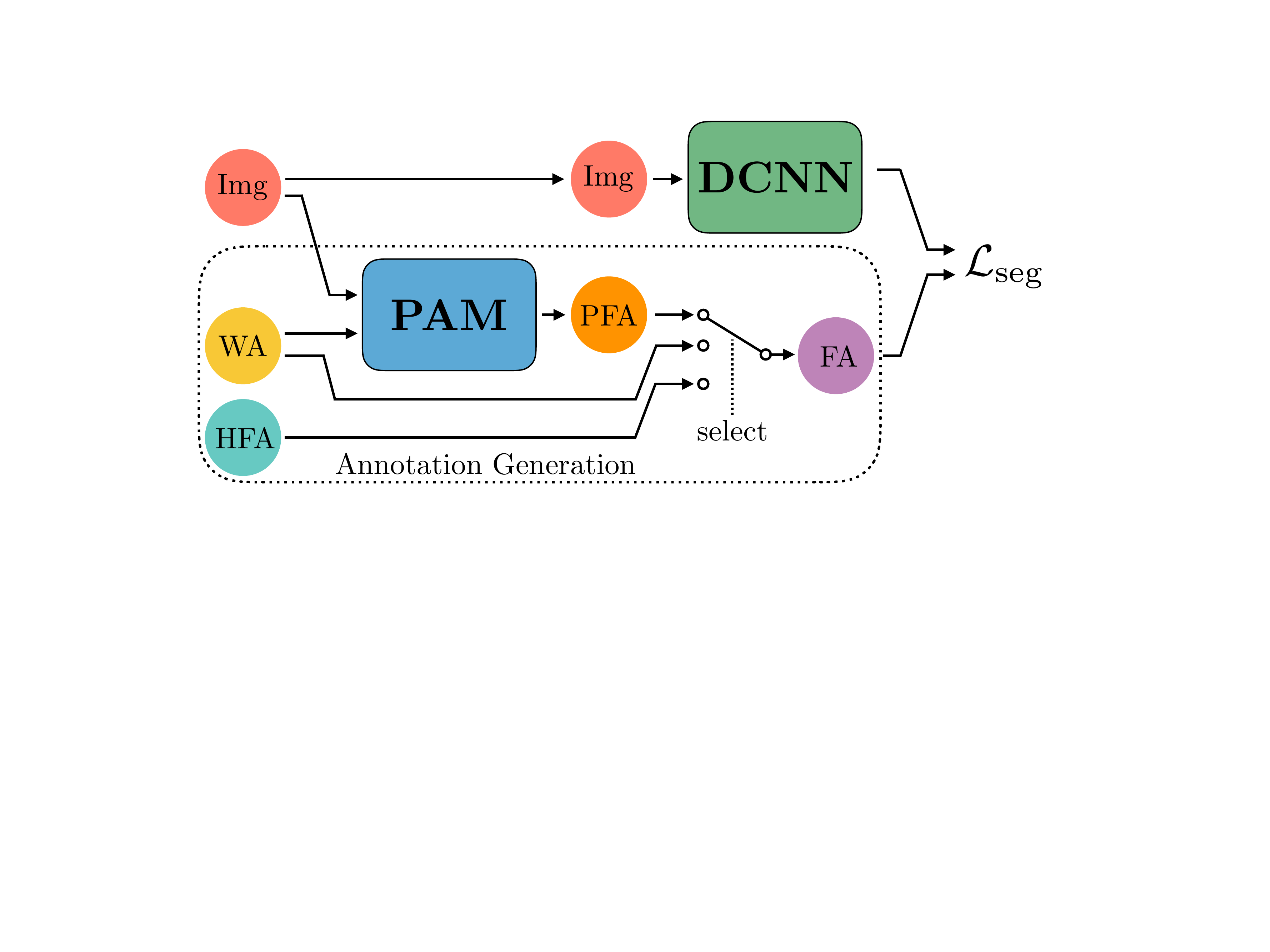}
  \caption{ %
    \textbf{Training a Segmentation \gls{dcnn} in a fully- and weakly-supervised
      scenario.} The segmentation cost $\mathcal{L}_\mathrm{seg}$ requires both
    training images (Img) and \acrlongpl{fa} (\acrshort{fa}). In a \emph{fully
      supervised scenario}, \acrshortpl{fa} are produced by humans
    (\acrshort{hfa}: Human \acrshort{fa}). In a \emph{weakly supervised
      scenario}, humans produce only weak annotations (WA). WAs can be used
    directly for training as \acrshortpl{fa} by using a special label at the
    unlabelled pixels, \eg \texttt{None} and define accordingly
    $\mathcal{L}_\mathrm{seg}$ to be 0 at these pixels. \acrshortpl{wa} can also
    be used to train an intermediate \acrlong{pam} (\acrshort{pam}) that will
    predict the classes for the missing annotations. \acrshortpl{pam} can be
    trained independently for each training image (local \acrshort{pam}), or
    trained on the whole training set (global \acrshort{pam}).}
    \label{fig:pam}
\end{figure}

%%% Local Variables:
%%% mode: latex
%%% TeX-master: "egpaper"
%%% End:

\subsection{Overview of the experimental strategy}

\paragraph{Weak supervision strategies.}
\label{sec:strategies}
Fig.~\ref{fig:pam} displays the different levels of training required in a fully
and a weakly supervised scenario. In both scenario, the training of the
segmentation \gls{dcnn} requires \glspl{fa}.
For full supervision, \glspl{fa} are directly provided by humans, hence called
\glspl{hfa}.

For weak supervision, \glspl{wa} need to be extended to \glspl{fa}. This can be
achieved in two ways.
The simplest way is to introduce an additional label that is used for unlabelled
pixels, and define the segmentation cost $\mathcal{L}_\mathrm{seg}$ at these
labels to be 0. This amounts to training only on the annotated pixels data. This
defines a \emph{baseline training strategy} because it corresponds to a default
strategy that does not attempt to predict the unlabelled pixels and that makes
the minimum modifications to the segmentation cost.

Another strategy amounts to use the \glspl{wa} to train an intermediate model to
predict the semantic classes at the unlabelled locations, resulting in a
\gls{pfa}. This is called a \acrlong{pam} (\acrshort{pam}).

\paragraph{Defining the Gap between full and weak supervision.}
Our goal is to unravel the ingredients required to train a \gls{dcnn} on
\glspl{wa} that achieves performances comparable to \emph{the same} \gls{dcnn}
trained on \glspl{hfa}.
We define the gap as the segmentation accuracy difference between the network
trained in the full and in the weak supervision scenario. If we use \gls{miou}
as the accuracy measure, this writes:
\begin{equation}
  \mathrm{Gap} \defeq \mathrm{mIoU}^\mathrm{Full}-\mathrm{mIoU}^\mathrm{Weak}\ .
\end{equation}

We assess a given training strategy by computing the relative reduction of the
Gap compared to the baseline strategy.

\paragraph{Assessing \acrshortpl{pam}.}
\Glspl{pam} introduce an intermediate level of training.
We evaluate their prediction quality by reporting the predicted annotations
accuracy (column \gls{pfa} in Tab.~\ref{table:summary}).
In addition, we also report the accuracy of the \gls{dcnn} after training
(sub-column Train below \gls{dcnn} in Tab.~\ref{table:summary}).
This is usually omitted in the literature tackling the full supervision problem
because the predictions of the trained network on the training set will
automatically be worse than the original annotations.

However it is unclear how the accuracy of the predicted annotations will compare
to the predictions of the network after training because they correspond to two
levels of training.
In addition, this comparison is important to study how improvements in the
predicted annotation accuracy (measured on the training set) translate into
improvements of the segmentation accuracy of the resulting trained network
(measured on the validation set).

\subsection{Data and weak annotations}
\label{sec:data-weak-annot}

Following~\cite{Lin2016,Tang2018,Tang2018ECCV}, we use the PASCAL VOC data
set~\cite{Everingham2010} with the publicly available scribble
\glspl{wa}~\cite{Lin2016}.

\paragraph{Curated \acrshortpl{wa}: assessing human annotation errors.} 
Inevitably, weak annotations contain errors.
For scribbles, two sources are possible: assigning the wrong class or annotating
multiple classes with one stroke.
To assess the impact of human annotation errors in the weakly supervised
scenario, we assemble a curated training set.
We keep the scribble positions but we assign the ground truth semantic class to
each annotated pixel. Furthermore, we require that, for each image, the human
annotator has labelled all the classes present in the ground truth.
We drop the images not satisfying this requirement: the curated training set
contains 10489 instead of 10582 images.

We always report the results about the best strategy for both the curated and
the original scribbles (Tab.~\ref{table:summary}, lines 8/9 and 12/13,
Tab.~\ref{table:comparison}).
However, for the sake of simplicity, we show the results about the different
ingredients in the training strategy that lead us to identifying our best
strategy only on the curated dataset (Tab.~\ref{table:summary}, lines 3--9 and
Fig.~\ref{fig:overview}).

\subsection{Predictive annotator models}
\label{sec:pred-models}

\Glspl{pam} can be trained for each training image independently (many
\emph{local \glspl{pam}}) or for the whole training set (one single \emph{global
  \gls{pam}}).

\subsubsection{Local \Gls{pam}: random forest}
\label{sec:rf}
In interactive image segmentation two \glspl{pam} are common:
\glspl{gmm}~\cite{Rother2004a,Nieuwenhuis2013a} and \gls{rf}
classifiers~\cite{Breiman2001,Santner2011}.
We use \gls{rf} instead of \gls{gmm} because \glspl{rf} train quickly even with
high dimensional data and provides a feature importance score enabling feature
ranking.
We use the custom Gini \emph{feature importance score} available in python in
\verb|scikit-learn|.
The features found in \glspl{dcnn} trained for classification display desirable
properties such as compositionality, invariance and class discrimination for
ascending layers~\cite{Zeiler2014}.
The first-layer contain low-level image filters for different colour patterns
\eg edge or corner filters.
Thus, we propose using the first-layer features of VGG-16~\cite{Simonyan2015}
and AlexNet~\cite{Krizhevsky2012} to train the \gls{rf} classifier on each
training image.
We use 50 trees and we apply the feature importance score to select the 100 most
informative ones out of 160.
Santner \emph{et al.}~\cite{Santner2011,Santner2009} use 30, 100 and 250 trees.
We have observed that using 50 trees and 100 features leads to finer-grained
predictions and decreases the amount of strong false positives and false
negatives, \ie class predictions with probability estimates of 0.0 or 1.0,
preventing further improvement by regularisation.

\subsubsection{Global \Gls{pam}: DeepLab}
\label{sec:glob-glsplp-deepl}

A state-of-the-art \gls{dcnn} in supervised semantic segmentation is
DeepLab~\cite{Chen2016b}.
Different versions of DeepLab are used in weakly-, semi-, or fully-supervised
settings~\cite{Lin2016,Papandreou}. We propose to use a simplified version of
DeepLabV2 as the global \gls{pam}.
We avoid pre-training DeepLabV2 on any segmentation dataset such as
\acrshort{ms-coco}~\cite{Chen2016b}.
For a detailed description of the exact architecture, training strategy and
hyper parameter optimization, we refer the reader to the supplementary material.

\subsection{Regularising \Glspl{pam}}
\label{sec:variational-models}

A Bayesian semantic segmentation model consists of a \emph{semantic model}
encoding how well pixels fit in their putative classes (encoded in a
\emph{data-fitting term}) and a consistency model encoding how well a particular
segmentation fits some desired prior knowledge (encoded in a
\emph{regularisation term}).
User inputs are used to train the semantic model derived from the \glspl{pam} (
described in Sec.~\ref{sec:pred-models}) that learn how pixels should be
labelled according to the human annotator.
Regularisation helps when the semantic model is uncertain about how to assign a
class to a pixel, in particular when the amount of pixels to calibrate the
semantic model is low.
In this work, we compare two popular variational models:
Potts~\cite{Nieuwenhuis2013} and \gls{fc-crf}~\cite{Krahenbuhl2011}. Both share
the same data-fitting term, but differ in their regularisation.

\paragraph{Bayesian data-fitting terms.} 
The data-fitting term writes as the sum over the pixels (denoted $\pixelSet$) of
the scalar product between the semantic \emph{segmentation mask} vector, denoted
$\mask_\pixel$, and the negative $\log-$\emph{labelling probability} vector,
denoted $\classFreq_\pixel$, see for example~\cite{Nieuwenhuis2013}:
\begin{equation}
  \label{eq:general-data-fitting-term}
  \Edf(\classFreq, \mask) \defeq
  \sum_{\pixel \in \pixelSet} \inner{-\log \classFreq_\pixel}{\mask_\pixel}~.
\end{equation}
The vector $\classFreq_\pixel$ encodes the semantic segmentation model:
$\classFreq_{\pc}$ is the probability of assigning class $\class \in \classSet$
at pixel $\pixel$.
The vector $\mask_\pixel$ represents a valid semantic segmentation hypothesis,
\ie an element of the unit probability simplex.
A mask vector containing only zeroes and ones corresponds to a proper labelling,
\ie a unique label is assigned everywhere.
Otherwise, the segmentation is called soft, and a proper labelling is recovered
by selecting at each pixel $\pixel$ the class with highest value in
$\mask_\pixel$.

The probability vector $\classFreq$ is derived from the \glspl{pam} (see
Sec.~\ref{sec:pred-models}) by normalising appropriately their soft predictions
in the probability simplex. They are denoted $\classFreq^\mathrm{local}$ and
$\classFreq^\mathrm{global}$ for the local and global \glspl{pam} respectively.

\paragraph{Potts regularisation.}
The Potts model penalises the total length of the interface between classes. It
is popular for its simplicity and the various efficient algorithms to compute
the associated \gls{map} (see \cite{Nieuwenhuis2013}).
This regularisation amounts to summing the weighted total variation of the mask
for each class:
\begin{equation}
  \label{eq:TV-regularisation-term} 
  \ErTV(\mask) \defeq \regp \sum_{\class \in \classSet} \TVgf(\mask_\class)\ ,
\end{equation}
where $g$ is the edge-stop function driving the class boundaries towards high
image gradients. It is defined as
$g_\pixel = \exp(-\eta \abs{\grad{\image}}_\pixel)$.

\paragraph{\acrshort{fc-crf} regularisation.}
For comparison, we also use the more complex regularisation of the \gls{fc-crf}
model~\cite{Krahenbuhl2011}.
It is popular in the literature as a post-processing to recover detailed
features from the blobby predictions of a \gls{dcnn}.
This regulariser is defined as
\begin{equation}
  \label{eq:FCCRF-regularisation-term}
  \ErCRFf(\mask) \defeq \sum_{\mathclap{\class \in \classSet, (\pixel, \pixelOther) \in \pixelSet^2}} [\mask_\pixel \neq \mask_\pixelOther] \left( w_1\, \appearanceKernel_{\pixel\pixelOther}+w_2\, \smoothnessKernel_{\pixel\pixelOther} \right)\ ,
\end{equation}
where
$\appearanceKernel_{\pixel\pixelOther} \defeq
\exp\left(-d_{\pixel\pixelOther}^2/2\sigma_\alpha^2-\delta_{\pixel\pixelOther}^2/2\sigma_\beta^2\right)$
and
$\smoothnessKernel_{\pixel\pixelOther} \defeq
\exp\left(-d_{\pixel\pixelOther}^2/2\sigma_\gamma^2\right)$ are the
\emph{appearance} and \emph{smoothness} kernel respectively, with
$d_{\pixel\pixelOther}$ the Euclidean distance between pixels $\pixel$ and
$\pixelOther$, and $\delta_{\pixel\pixelOther}$ the Euclidean distance in color
space between pixels $\pixel$ and $\pixelOther$.
This prior tends to group nearby pixels with similar colours (appearance
kernel), and at the same time penalise small clusters (smoothness kernel).

\paragraph{\acrshort{map} annotations prediction.} 
For both models we solve the associated \gls{map} problem
\begin{equation}
  \label{eq:MAP}
  \mathrm{M}^\textrm{MAP}(\classFreq) \defeq \arginf_{\mask \in \simplex} \Edf(\classFreq, \mask)+\Erf(\mask)\ ,
\end{equation}
with $\simplex \defeq \bigcup \simplex_\pixel\ $ the set of valid soft
segmentation masks, and $\simplex_\pixel$ the probability simplex defined at
pixel $\pixel$.
The \gls{map} problem aims at finding the optimal trade-off between fitting the
data, driven by $\Ed$ (see Eq.~\eqref{eq:general-data-fitting-term}) and the
regularisation $\Er$, being either Potts (see
Eq.~\eqref{eq:TV-regularisation-term}) or \gls{fc-crf} (see
Eq.~\eqref{eq:FCCRF-regularisation-term}).

We solve the Potts \gls{map} problem using the strategy described
in~\cite{Paul2011}.
For the \gls{fc-crf} \gls{map} problem, we use the implementation
of~\cite{Krahenbuhl2011} found at~\cite{Beyer2018}.
We refer the reader to the supplementary material for a detailed account of the
procedure to select the regularisation parameters and the model-dependent
parameters.

We denote by
$\mask^\textrm{local} \defeq \mathrm{M}^\textrm{MAP}(\classFreq^\textrm{local})$
and
$\mask^\mathrm{global} \defeq
\mathrm{M}^\mathrm{MAP}(\classFreq^\mathrm{local})$ the regularisation of the
local and the global \glspl{pam} respectively.

\subsection{Predicted annotations from \Glspl{pam}}
For each training image, the \acrlongpl{pfa} are obtained at each pixel $\pixel$
independently from a soft segmentation $\vec{S} \in \simplex$ by:
\begin{equation}
  \label{eq:pfa}
  \mathrm{PFA}_\pixel(\vec{S}) \defeq \argmax_{\class \in \classSet} \vec{S}_{\pixel}\ .
\end{equation}
Without regularisation, we define the local and the global \glspl{pfa} by
\begin{align}
  \mathrm{PFA}^\mathrm{local}  & \defeq \mathrm{PFA}(\classFreq^\mathrm{local}) \label{eq:pfa-local}\\ %\text{and}\ 
  \mathrm{PFA}^\mathrm{global} & \defeq \mathrm{PFA}(\classFreq^\mathrm{global})\label{eq:pfa-global}\ .
\end{align}
We define the regularised \glspl{pfa} by
\begin{align}
  \mathrm{PFA}^\mathrm{local+MAP}  & \defeq \mathrm{PFA}(\mask^\mathrm{local}) \label{eq:pfa-map-local} \\% \text{and}\ 
  \mathrm{PFA}^\mathrm{global+MAP} & \defeq \mathrm{PFA}(\mask^\mathrm{global})\label{eq:pfa-map-global} \ .
\end{align}

%%% Local Variables:
%%% mode: latex
%%% TeX-master: "egpaper"
%%% End:

%  LocalWords:  Variational componentwise rescaling rescaled blobby regulariser
%  LocalWords:  RGB variational functionals reproducibility bilinear upsampling
%  LocalWords:  dataset

\section{Experiments and results}
\begin{table*}[!t]
  \renewcommand{\arraystretch}{1.2}
  \newcommand{\head}[1]{\textnormal{\textbf{#1}}}
  \newcommand{\normal}[1]{\multicolumn{1}{c}{#1}}
  \newcommand{\na}{\emph{n.a.}}

  \colorlet{tableheadcolor}{gray!25} % Table header colour = 25% gray
  \newcommand{\headcol}{\rowcolor{tableheadcolor}} %
  \colorlet{tablerowcolor}{gray!10} % Table row separator colour = 10% gray
  \newcommand{\rowcol}{\rowcolor{tablerowcolor}} %
  \setlength{\aboverulesep}{0pt}
  \setlength{\belowrulesep}{0pt}

  \newcommand*{\rulefiller}{
    \arrayrulecolor{tableheadcolor}% change to cell colour
    \specialrule{\heavyrulewidth}{0pt}{-\heavyrulewidth}% "invisible" rule
    \arrayrulecolor{black}}% revert to regular line colour

  \newcounter{magicrownumbers}
  \newcommand\rownumber{\stepcounter{magicrownumbers}{\tiny \arabic{magicrownumbers}}}
  
    \centering
    \caption{\textbf{Towards closing the gap}. All the values correspond to
      \acrshort{miou} [\%] except the last column. The \textit{Full Gap}
      corresponds to the gap between the two methods shown in light grey, for
      both the curated and the original \acrshortpl{wa}~\cite{Lin2016}. The
      \textit{Remaining Gap} defines the difference between DeepLabV2 trained on
      human or predicted full annotations. \textit{Gap Reduction (\%)} is
      defined by
      $(\textrm{Full Gap} - \textrm{Remaining Gap})/\textrm{Full Gap}$. The
      colored symbols correspond to Fig.~\ref{fig:overview}. The column
      \acrshort{pfa} reports the \acrshort{miou} of the \acrshortpl{pfa} of the
      training images produced by the \acrshortpl{pam}. }
    \label{table:summary}

  \resizebox{0.95\textwidth}{!}{%
  \begin{tabular}{*{6}{c}r*{6}{c}}
    \toprule%[1.5pt]

    \headcol \multicolumn{2}{c}{\head{Training Data}} & \multicolumn{2}{c}{\head{PAM}}
    & \multicolumn{2}{c}{\head{Regularisation}} & \head{Marker} &\head{PFA} & \multicolumn{2}{c}{\head{DCNN}} & \multicolumn{3}{c}{\head{Gap}} \\
    
    \headcol &                       & Local      & Global     & Potts      & FC-CRF     &  Fig.~\ref{fig:overview} & Train & Train & Val & Full & Remaining & Reduction (\%)    \\ 
    \toprule
    \rowcol  & HFA                   &            &            &            &            &                  & 100   & 80.7  & 71.5 & \na          & \na          & \na       \\
    \rowcol  & WA                    &            &            &            &            & \soft{global}    & 100   & 74.8  & 67.1 & \na          & \na          & \na       \\	
             & \multirow{ 7}{*}{PFA} & \checkmark &            &            &            & \soft{local}     & 61.5  & 67.8  & 61.0 & 4.4          & 10.5         & -138.6\%  \\
             &                       & \checkmark &            & \checkmark &            & \potts{local}    & 73.3  & 70.9  & 63.3 & 4.4          & 8.2          & -70.4\%   \\
             &                       &            & \checkmark & \checkmark &            & \potts{global}   & 79.4  & 74.6  & 67.8 & 4.4          & 3.7          & 15.9\%    \\
             &                       &            & \checkmark &            & \checkmark & \fccrf{global}   & 79.6  & 75.6  & 68.2 & 4.4          & 3.3          & 25.0\%    \\
             &                       & \checkmark & \checkmark &            &            & \soft{combined}  & 81.0  & 76.4  & 69.2 & 4.4          & 2.3          & 47.7\%    \\
             &                       & \checkmark & \checkmark & \checkmark &            & \potts{combined} & 84.2  & 76.8  & 69.7 & 4.4          & 1.8          & 59.1\%    \\
    \parbox[t]{2mm}{\multirow{-8}{*}{\rotatebox[origin=c]{90}{Curated}}}
             &                       & \checkmark & \checkmark &            & \checkmark & \fccrf{combined} & 83.9  & 77.6  & 70.0 & 4.4          & \textbf{1.5} & \textbf{65.9\%} \\
    \toprule
    \rowcol  & HFA                   &            &            &            &            & & 100   & 80.7   & 71.5  & \na          & \na          & \na    \\
    \rowcol  & WA                    &            &            &            &            & & 95.4  & 71.7   & 64.3  & \na          & \na          & \na    \\
             & \multirow{ 2}{*}{PFA} & \checkmark & \checkmark & \checkmark &            & & 81.6  & 76.4   & 68.8  & 7.2          & 2.7          & 62.5\% \\
    \parbox[t]{2mm}{\multirow{-4}{*}{\rotatebox[origin=c]{90}{Original}}}
             &                       & \checkmark & \checkmark &            & \checkmark & & 81.4  & 76.5   & 69.1  & 7.2          & \textbf{2.4} & \textbf{66.7\%} \\
    \bottomrule%[1.5pt]
  \end{tabular}
  }
\end{table*}
%%% Local Variables:
%%% mode: latex
%%% TeX-master: "egpaper"
%%% End:

To explore and identify weak supervision strategies we use DeepLabV2 both as the
global annotator model and as the segmentation \gls{dcnn} (see
Fig.~\ref{fig:pam} and Sec.~\ref{sec:strategies}).
However, after having identified the best strategy, we test different networks
and architectures for the segmentation \gls{dcnn} in Sec.~\ref{sec:comparison}.
We report the corresponding values in Tab.~\ref{table:summary},
\ref{table:comparison} and Fig.~\ref{fig:overview}.

\subsection{Measuring the Gap}

Under full supervision, DeepLabV2 achieves a \gls{miou} of 71.5 on the
validation set, which is consistent with the performances reported for the
TensorFlow implementation~\cite{Wang2018, Wang2018smoothed}.

Under weak supervision (see Sec.~\ref{sec:strategies}), DeepLabV2 trained solely
on the original scribbles found in~\cite{Lin2016} achieves a \gls{miou} of 64.3
on the validation set, establishing a gap of 7.2.
As expected, on the curated annotations, the \gls{miou} is higher (67.1) , and
the gap smaller (4.4), see Tab.~\ref{table:summary}.
This gives a first \emph{insight on the adversarial impact of the errors in the
  \gls{wa} on the resulting trained model}.

We explore different weak supervision strategies on the curated annotations
because the gap is smaller due to a baseline model that has already a high
\gls{miou} (67.1).
This makes it harder for any strategy to be elected.
However, for our best weak supervision strategy, we always report the results on
the original scribbles dataset, because in practice curated annotations do not
exist, see Tab.~\ref{table:summary}.
In addition, when comparing our results to others in
Tab.~\ref{table:comparison}, we always use the original scribble dataset of
\cite{Lin2016}.

\subsection{Local \Glspl{pfa} are worse than the baseline}
\begin{figure}[!h]
  \centering
  \includegraphics[width=\columnwidth]{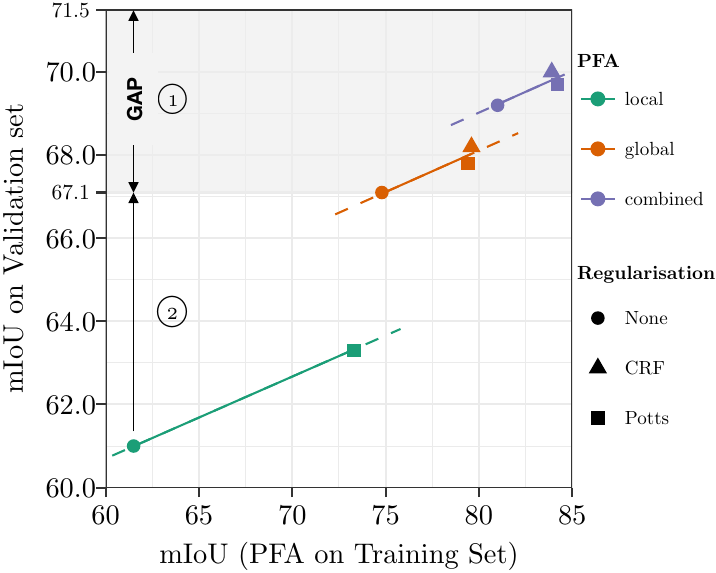}
  \caption{\textbf{Visual summary of our study}. The gap is the difference
    between DeepLabV2 trained solely on scribbles and DeepLabV2 trained on the
    human full annotations. See Tab.~\ref{table:summary}.}
  \label{fig:overview}
\end{figure}

%%% Local Variables:
%%% mode: latex
%%% TeX-master: "egpaper"
%%% End:

\label{sec:assess-qual-glsplpf}

We recall that the global annotator model, \ie the baseline, is DeepLabV2
trained solely on the scribbles (see Sec.~\ref{sec:strategies}). This represents
only 1.6\% of pixels annotated on average per class. We expect that annotating
all pixels by training a local annotator model per training image helps. However
these additional annotations are predicted, and hence prone to errors.

\paragraph{Assessing the local \glspl{pfa}.}
On each training image we train a \gls{rf} (see Sec.~\ref{sec:rf}) on the
scribbled pixels, and we use it to predict (using Eq.~\ref{eq:pfa-local}) the
class for the unlabelled ones.
These predicted annotations achieve a \gls{miou} of 61.5, significantly lower
than the baseline that achieves annotations with a \gls{miou} of 74.8.
Potts regularisation for each local annotator (Eq.~\ref{eq:pfa-map-local})
improves the \gls{miou} of the predicted annotations by 11.8, up to 73.3.
It lags behind the baseline only by 1.1.
We interpret this result by the fact that the global \gls{pam} is trained on all
the scribbles, whereas each local \gls{pam} is trained only on the scribbles
available for the image where it predicts the missing annotations.
To some extent, Potts regularisation compensates for this scarce amount of
training pixels.

\paragraph{The relationship between \glspl{pfa} quality and the accuracy after
  training is nonlinear.}
A priori, we expect that the better the predicted annotations, the better the
segmentation after training.
We also expect that for a given improvement in the predicted annotations
quality, we obtain a lesser improvement after training of the resulting network.

After training DeepLabV2 on the local \glspl{pfa}, we obtain on the validation
set a \gls{miou} of 61.0 (without regularisation, \soft{local}) and 63.3 (with
regularisation, \potts{local}).
As expected, the improvement by 11.3 of the \gls{miou} due to regularisation of
the local \glspl{pfa}, yields after training a lesser improvement of 2.3 on the
validation set.
Surprisingly, the regularised local \glspl{pfa}, having an accuracy comparable
to the baseline, yield after training an accuracy 3.8 points below the baseline,
see Fig.~\ref{fig:overview}.

\begin{figure}
	\centering
	\def\svgwidth{1.0\columnwidth}
	{
    \fontsize{9pt}{11pt}\selectfont
    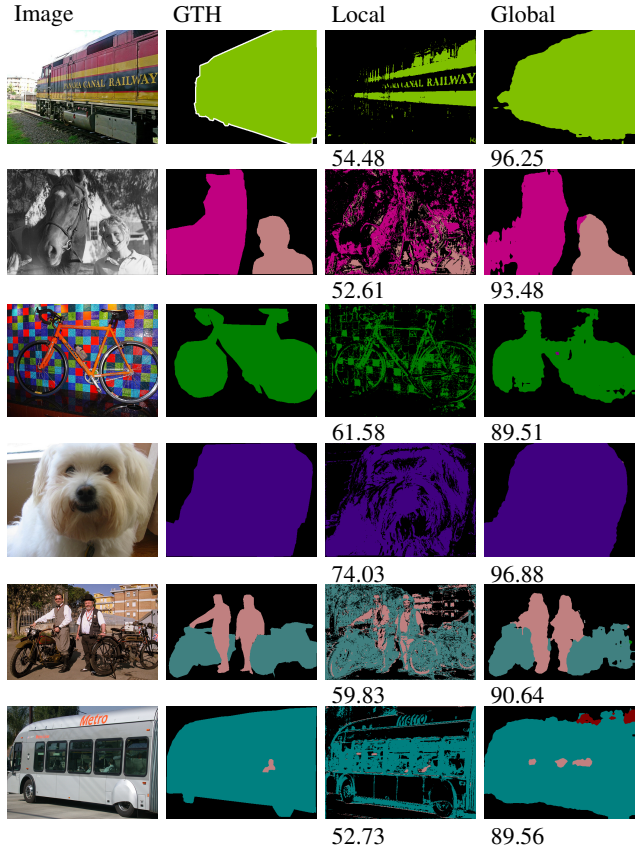
    }
    \caption{Examples where global \glspl{pfa} have a higher pixel accuracy than
      the local ones.}
    \label{fig:local-global-images}
\end{figure}

These findings suggest that the annotation errors predicted by the local and the
global \glspl{pam} are qualitatively different.
Fig.~\ref{fig:local-global-images} shows examples of images where the local
\gls{pam} is worse than the baseline.
In such cases, we observe that the local annotators produce scarce and
inconsistent labelling, whereas the global annotator yields overall good
annotations, but with imprecise boundaries.

\begin{figure}
  \centering
  \includegraphics{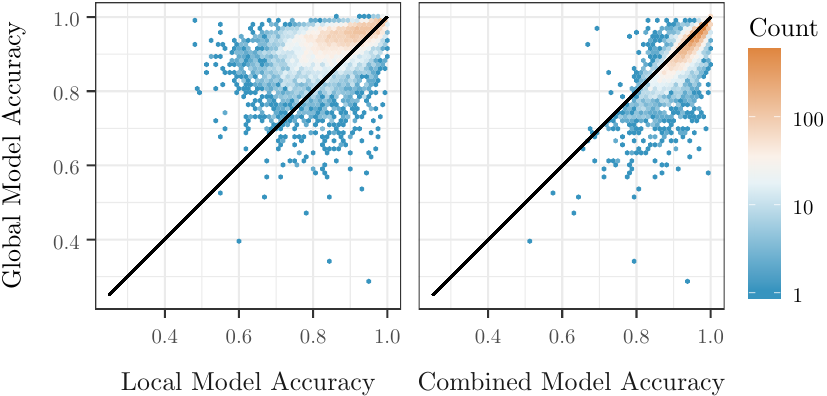}
  \caption{\textbf{Comparing local, global, and combined \glspl{pfa}.} The
    accuracy is the \gls{miou}. The diagonal is the equal accuracy cutoff.}
      \label{fig:hexbin}
\end{figure}

However, local models have the potential to capture better, in a given training
image, the labelling intended by the human annotator.
To quantify this effect we compare the pixel accuracy of the \glspl{pfa}
generated by the local and the global \glspl{pam}.
In Fig.~\ref{fig:hexbin} (left panel), we observe that for most images, the
pixel accuracy is best for the baseline (\ie the global \gls{pam}), but a fair
amount of images are better labelled by the local \glspl{pam}.

%%% Local Variables:
%%% mode: latex
%%% TeX-master: "egpaper"
%%% End:

%%% Local Variables:
%%% mode: latex
%%% TeX-master: "egpaper"
%%% End:

\subsection{Regularising global \Glspl{pfa} improves only marginally}

In the previous section, we have learned that local \glspl{pam} are not good
enough to beat the baseline that is trained solely on scribbles.
But we have shown that regularising the predicted annotations can still
translate in a significant improvement in the resulting trained network.

Regularising the annotations predicted by the global \gls{pam} with Potts
(\potts{global}) boosts the \glspl{pfa} by 4.6 up to 79.4 in \gls{miou} and
leads to 67.8 after training.
Using the more complex \gls{fc-crf} regulariser (\fccrf{global}) leads to
comparable improvements in both the \glspl{pfa} quality (79.6) and the accuracy
of the trained network (68.2).

Regularising the annotation predictions by the baseline does help closing the
gap. However, the improvement is only marginal and reduces the gap only by
15.9\% (Potts) and 25.0\% (\gls{fc-crf}).

\begin{figure}
	\centering
	\def\svgwidth{1.\columnwidth}
	{\fontsize{9pt}{11pt}\selectfont
    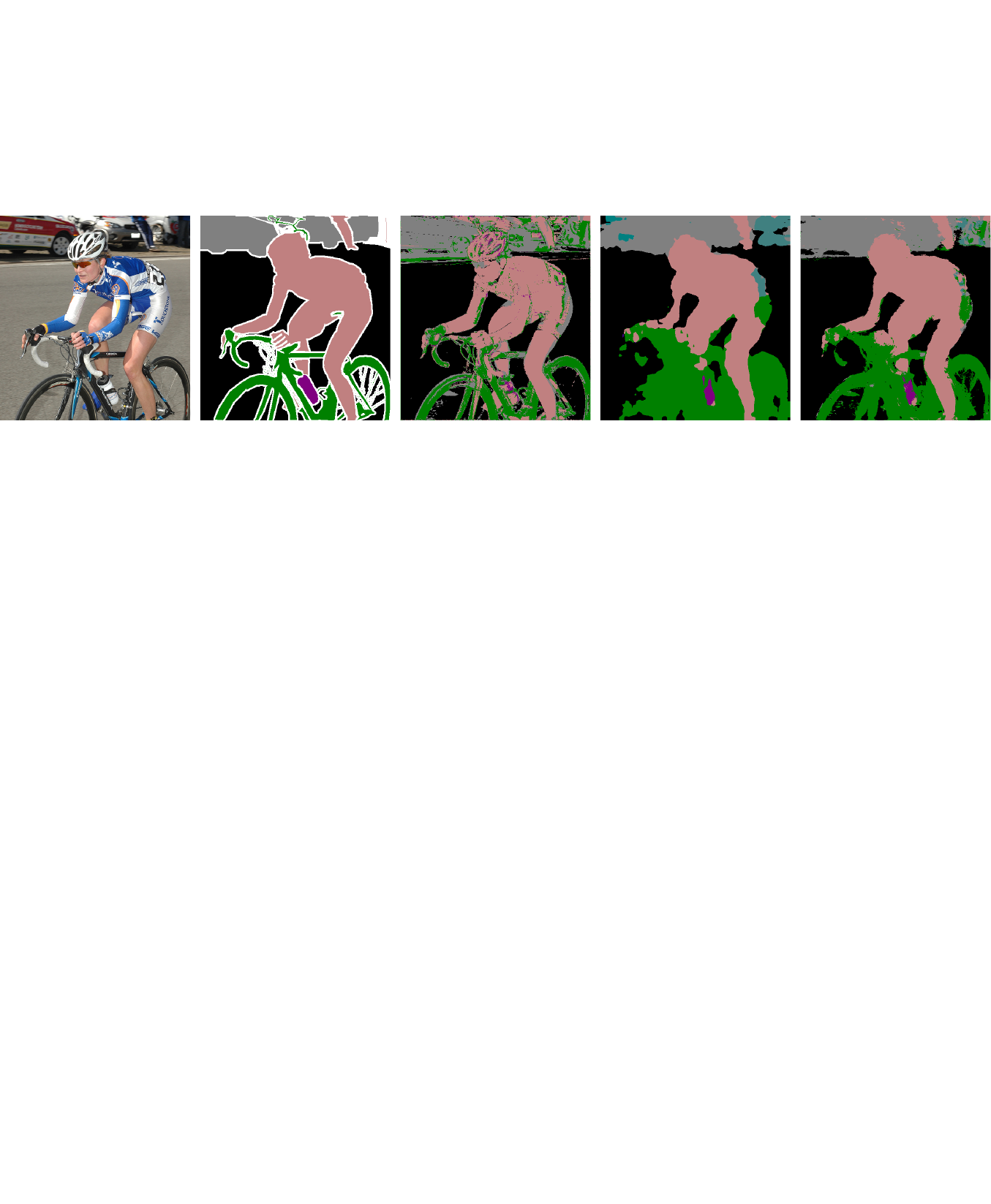
    }
    \caption{ Examples where the combined \glspl{pfa} have a higher accuracy
      than either the local or the global ones.}
    \label{fig:pfa-combined}
\end{figure}

%%% Local Variables:
%%% mode: latex
%%% TeX-master: "egpaper"
%%% End:

\subsection{Combining local and global \Glspl{pfa} is best}

In Sec.~\ref{sec:assess-qual-glsplpf} we have observed that there are images
where the local \glspl{pam} achieve better annotations (\ie below the diagonal
in Fig.~\ref{fig:hexbin}, left panel).
This suggests that the local and the global \glspl{pam} can be complementary.

To test this idea, we combine them by \emph{averaging their predictions}:
\begin{align}
  \classFreq^\mathrm{comb} & \defeq \textstyle\frac{1}{2} \classFreq^\mathrm{local}+\textstyle\frac{1}{2} \classFreq^\mathrm{global} \\
  \textrm{PFA}^\mathrm{comb} & \defeq \mathrm{PFA}(\classFreq^\mathrm{comb})\ .
\end{align}
The right panel of Fig.~\ref{fig:hexbin} shows that the combined local and
global annotation predictions improve overall compared to the global \gls{pam}
alone.

Fig.~\ref{fig:pfa-combined} shows images where the combined predictions have an
accuracy superior to either the local or the global \gls{pam}.
Furthermore, Fig.~\ref{fig:pfa-combined} shows examples where the global
\gls{pam} assigns class labels absent in the ground truth.
However, a local \gls{pam} avoid these errors because they are constrained to
predict, in a given training image, only the classes present in the scribble
annotations.
Therefore, combining local and global \glspl{pam} reduces the effect of
predicting foreign class labels.
These observations support our claim in Sec.~\ref{sec:assess-qual-glsplpf} that
the predicted annotations from the local and the global \glspl{pam} are
qualitatively different.

Without regularisation, $\textrm{PFA}^\textrm{comb}$ achieves a \gls{miou} of 81
(local: 61 and global: 74.8).
Subsequently training DeepLabV2 achieves 69.2 (\soft{combined}) on the
validation set.
Hence, combining local and global \glspl{pam} boosts the quality of the
\glspl{pfa} substantially.
The model trained on these improved \glspl{pfa} reduces the gap by 47.7\% down
to 2.3.

We use the averaged predictions $\classFreq^\textrm{comb}$ to obtain the
regularised combined predicted annotations :
\begin{equation}
  \textrm{PFA}^{\textrm{comb + MAP}} \defeq \mathrm{PFA}(\mathrm{M}^\textrm{MAP}(\classFreq^\mathrm{comb}))\ .
\end{equation}
As expected, regularisation improves both the \glspl{pfa} quality and the
accuracy after training, leading to models with a \gls{miou} of 69.7
(\potts{combined}) and 70.0 (\fccrf{combined}) on the validation set.
Hence, combining local and global \glspl{pam} by averaging their predictions has
a larger impact than regularisation alone.
However, in synergy, these two ingredients lead to the best models that reduce
the gap by 59\% to 1.8 (Potts) and 65.9\% to 1.5 (\gls{fc-crf}).

We make two additional controls for our best strategy: we test our weak
supervision strategy on the non-curated original scribbles of~\cite{Lin2016},
and on the test set of the PASCAL VOC2012 challenge.
On the non-curated annotations, we achieve even better results and reduce the
gap by 62.5\% (Potts) and 66.7\% (\gls{fc-crf}).
This shows that our strategy is robust to human annotation errors.
Additionally, we report the \gls{miou} on the test set, where we achieve 70.4
(\gls{fc-crf}).

\subsection{Comparison to others}
\label{sec:comparison}
\begin{table*}[t]
  \renewcommand{\arraystretch}{1.2}
  \newcommand{\head}[1]{\textnormal{\textbf{#1}}}
  \newcommand{\normal}[1]{\multicolumn{1}{c}{#1}}
  \newcommand{\na}{\emph{n.a.}}

  \colorlet{tableheadcolor}{gray!25} % Table header colour = 25% gray
  \newcommand{\headcol}{\rowcolor{tableheadcolor}} %
  \colorlet{tablerowcolor}{gray!10} % Table row separator colour = 10% gray
  \newcommand{\rowcol}{\rowcolor{tablerowcolor}} %
  \setlength{\aboverulesep}{0pt}
  \setlength{\belowrulesep}{0pt}

  \newcommand*{\rulefiller}{
    \arrayrulecolor{tableheadcolor}% change to cell colour
    \specialrule{\heavyrulewidth}{0pt}{-\heavyrulewidth}% "invisible" rule
    \arrayrulecolor{black}}% revert to regular line colour

    \centering
    \caption{\textbf{Comparison to state-of-the-art methods in weakly supervised
        semantic segmentation.} All the values correspond to \gls{miou} [\%] on
      the validation set of PASCAL VOC12. \textit{Weak + Baseline} corresponds
      to training on the raw \acrshortpl{wa}. \textit{Full} corresponds to
      training on the fully-annotated training set. \textit{Gap} is the
      difference between the two. \textit{Weak + Strategy} corresponds to
      training the \textit{Segmentation \gls{dcnn}} using the weak supervision
      strategy shown in column \textit{Strategy}. \textit{Remaining Gap}
      computes the difference between full and weak supervision.
      \textit{Reduction (\%)} is defined by
      $(\textrm{Full Gap} - \textrm{Remaining Gap})/\textrm{Full Gap}$.
      $(^{\ast})$~denotes methods using bounding box \glspl{wa}. Best values are
      highlighted in \textbf{bold}.}
    \resizebox{0.99\textwidth}{!}{%
      \begin{tabular}{llc*{8}{c}}
        \toprule
        \headcol \head{Strategy}                    & \head{Segmentation DCNN} & \head{MSC} & \head{CRF}       & \head{Weak + Baseline} & \head{Full} & \head{Gap} & \head{Weak + Strategy}    & \head{Remaining Gap} & \head{Reduction (\%)} \\
        \toprule
        \textbf{Ours}                            & DeepLab-ResNet101       & ---        & ---              & 64.3            & 71.5           & 7.2        & 69.1             & 2.4            & \textbf{66.7\%}   \\
        \textbf{Ours}                            & PSPNet-ResNet50         & ---        & ---              & 67.8            & 76.6           & 8.8        & 73.0             & 3.6            & 59.1\%            \\
        \textbf{Ours}                            & PSPNet-ResNet101        & ---        & ---              & 69.3             & 77.0          & 7.7        & \textbf{74.4}    & 2.6            & \textbf{66.2\%}               \\
        \midrule
        \textbf{Ours}                            & DeepLab-ResNet101       & ---        & \checkmark       & 66.4            & 73.4           & 7.0        & 70.5             & 2.9            & \textbf{58.6\%}   \\
        \textbf{Ours}                            & PSPNet-ResNet50         & ---        & \checkmark       & 69.2            & 77.2           & 8.0        & 73.5             & 3.7            & 53.8\%            \\
        \textbf{Ours}                            & PSPNet-ResNet101        & ---        & \checkmark       & 70.7            & 77.5           & 6.8        & \textbf{74.7}    & 2.8            & \textbf{58.8\%}               \\
        \midrule
        NCL~\cite{Tang2018}                      & DeepLab-VGG16           & \checkmark & ---              & 60.4            & 68.8           & 8.4        & 62.4             & 6.4            & 23.8\%            \\
        CRF~\cite{Tang2018ECCV}                  & DeepLab-VGG16           & \checkmark & ---              & 60.4            & 68.8           & 8.4        & 64.4             & 4.4            & 47.6\%            \\
        KernelCut~\cite{Tang2018ECCV}            & DeepLab-VGG16           & \checkmark & ---              & 60.4            & 68.8           & 8.4        & 64.8             & 4.0            & 52.4\%            \\
        \textbf{Ours}                            & DeepLab-ResNet101       & \checkmark & ---              & 65.6            & 73.7           & 8.1        & 70.8             & 2.9            & \textbf{64.2}\%   \\
        NCL~\cite{Tang2018}                      & DeepLab-ResNet101       & \checkmark & ---              & 69.5            & 75.6           & 6.1        & 72.8             & 2.8            & 54.1\%            \\
        CRF~\cite{Tang2018ECCV}                  & DeepLab-ResNet101       & \checkmark & ---              & 69.5            & 75.6           & 6.1        & 72.9             & 2.7            & 55.7\%            \\
        KernelCut~\cite{Tang2018ECCV}            & DeepLab-ResNet101       & \checkmark & ---              & 69.5            & 75.6           & 6.1        & 73.0             & 2.6            & 57.4\%            \\
        \textbf{Ours}                            & PSPNet-ResNet50         & \checkmark & ---              & 69.5            & 77.6           & 8.1        & 74.5             & 3.1            & 62.7\%            \\
        \textbf{Ours}                            & PSPNet-ResNet101        & \checkmark & ---              & 71.3            & 79.2           & 7.9        & \textbf{75.6}    & 3.6            & 54.4\%               \\
        \midrule
        ScribbleSup~\cite{Lin2016}               & DeepLab-LargeFOV        & \checkmark & \checkmark       & ---             & 68.7           & ---        & 63.1             & 5.6            & ---               \\ 
        SimpleDoesIt$^{\ast}$~\cite{Khoreva2017} & DeepLab-LargeFOV        & \checkmark & \checkmark       & ---             & 69.1           & ---        & 65.7             & 3.4            & ---               \\
        NCL~\cite{Tang2018}                      & DeepLab-VGG16           & \checkmark & \checkmark       & 64.3            & 71.5           & 7.2        & 65.2             & 6.3            & 12.5\%            \\
        CRF~\cite{Tang2018ECCV}                  & DeepLab-VGG16           & \checkmark & \checkmark       & 64.3            & 71.5           & 7.2        & 66.4             & 5.1            & 29.2\%            \\
        KernelCut~\cite{Tang2018ECCV}            & DeepLab-VGG16           & \checkmark & \checkmark       & 64.3            & 71.5           & 7.2        & 66.7             & 4.8            & 33.3\%            \\
        SimpleDoesIt$^{\ast}$~\cite{Khoreva2017} & DeepLab-ResNet101       & \checkmark & \checkmark       & ---             & 74.5           & ---        & 69.4             & 5.1            & ---               \\
        \textbf{Ours}                            & DeepLab-ResNet101       & \checkmark & \checkmark       & 67.6            & 75.1           & 7.5        & 72.1             & 3.0            & \textbf{60.0\%}   \\
        NCL~\cite{Tang2018}                      & DeepLab-ResNet101       & \checkmark & \checkmark       & 72.8            & 76.8           & 4.0        & 74.5             & 2.3            & 42.5\%            \\
        CRF~\cite{Tang2018ECCV}                  & DeepLab-ResNet101       & \checkmark & \checkmark       & 72.8            & 76.8           & 4.0        & 75.0             & 1.8            & 55.0\%            \\
        KernelCut~\cite{Tang2018ECCV}            & DeepLab-ResNet101       & \checkmark & \checkmark       & 72.8            & 76.8           & 4.0        & 75.0             & 1.8            & 55.0\%            \\
        \textbf{Ours}                            & PSPNet-ResNet50         & \checkmark & \checkmark       & 69.9            & 77.8           & 7.9        & 74.6             & 3.2            & \textbf{59.5\%}   \\
        \textbf{Ours}                            & PSPNet-ResNet101        & \checkmark & \checkmark       & 71.8             & 79.4          & 7.6        & \textbf{75.7}    & 3.7            & 51.3\%               \\
    \bottomrule%[1.5pt]
  \end{tabular}
  }
  \label{table:comparison}

\end{table*}

%%% Local Variables:
%%% mode: latex
%%% TeX-master: "egpaper"
%%% End: 

We compare our best weak supervision strategy identified in the previous
section.
To generate full annotations, we regularise with \gls{fc-crf} the averaged
predictions of the local (\glspl{rf}) and the global (DeepLabV2 without
multi-scaling) \glspl{pam}.
We use these predicted full annotations to train different \glspl{dcnn}:
DeepLabv2-ResNet101~\cite{Chen2016b}, PSPNet-ResNet50 and
PSPNet-ResNet101~\cite{Zhao2017,Li2018} (see the supplementary material for
details).
We also test popular improvement strategies such as post-processing (CRF column
in Tab.~\ref{table:comparison}) and multi-scaling (MSC column in
Tab.~\ref{table:comparison}).
In Tab.~\ref{table:comparison}, we report the comparison results on the
validation dataset of PASCAL VOC12.
We achieve \emph{new state-of-the-art performances in weakly supervised image
  segmentation with pixel-based annotations} with a \gls{miou} of 75.6 without
and 75.7 with post-processing.

ScribbleSup~\cite{Lin2016} introduced a regularisation strategy for the
predicted annotations based on superpixels and \acrshortpl{crf}.
In our framework, this corresponds to a regularised global \gls{pam} supervision
strategy.
However, their main improvement comes from iteratively retraining DeepLab, which
is computationally expensive.
Avoiding iterative retraining, our weak supervision strategy is faster and
achieves lower gaps (3.0--3.7 \vs 5.6) and higher absolute values (75.7 \vs
63.1).

SimpleDoesIt~\cite{Khoreva2017} uses another pixel-based weak annotation
modality: bounding boxes.
However, their strategy is similar to ours in spirit.
They designed a variant of GrabCut to predict \emph{partial} annotations and
segmentation proposal techniques to train DeepLab.
We achieve a lower gap (3.7 vs. 5.1) and a higher \gls{miou} (75.7 vs. 69.4).

Recently, NCL~\cite{Tang2018}, CRF~\cite{Tang2018ECCV},
KernelCut~\cite{Tang2018ECCV} established new state-of-the-art results by using
traditional variational segmentation models within a deep learning framework to
enable end-to-end training.
These methods require implementing graph-cut algorithms within deep learning
frameworks~\cite{Tang2019pytorch,Tang2019caffe} via new special layers that
increase considerably the training time.
Our strategy is more versatile because it does not require modifying the
architecture of \glspl{dcnn}.
Our weak supervision strategy achieves higher absolute values (75.7 vs. 75.0)
and reduces the gaps by a higher relative value (with MSC and without
\acrshort{crf}: 64.2\% (ours), 54.1\% (NCL), 55.7\% (CRF), 57.4\% (KernelCut)).

%%% Local Variables:
%%% mode: latex
%%% TeX-master: "egpaper"
%%% End:
%  LocalWords:  regulariser

\section{Conclusions and discussion} 

Our study tackles the challenging problem of training semantic segmentation
\glspl{dcnn} in a weakly-supervised setting.
We establish new experimental standards for this problem (see
Tab.~\ref{table:summary} and Fig.~\ref{fig:overview}): measuring the gap by
training solely on the weak annotations \textcircled{\tiny{1}}, quantifying the
adversarial effect of annotation errors \textcircled{\tiny{2}}, and comparing
different annotator models.
This allows us to unravel a counter-intuitive finding: \textit{averaging poor
  local predicted annotations with the baseline ones and reuse them for training
  a \gls{dcnn} yields new state-of-the-art results}
(\soft{combined},\potts{combined},\fccrf{combined}).
The fact that we achieve this without resorting to new cost-functions,
regularization or architectures is a strength: it allows others to adapt our
ideas to their setting and to extend the spectrum of strategies available for
weakly-supervising \glspl{dcnn}.

\paragraph{Motivations.} This study started from two observations:
\begin{itemize}
\item DeepLabV2 trained solely on weak annotations (\soft{global}) already beats
  custom strategies like ScribbleSup.
\item Additional annotations (62.5 times more for our data set) predicted by a
  model contain unavoidably errors. They have an adversarial effect on the
  resulting trained segmentation network (\soft{local},\potts{local}).
\end{itemize}

\paragraph{Insights.} Our study allows tackling these questions:
\begin{itemize}
\item \emph{Have all the errors in the predicted annotations the same
    adversarial effect?}
  No. After regularisation the local PFAs have a mIoU comparable to the
  baseline, but after training, the errors from the local PFAs have a higher
  negative impact on the accuracy on the validation set (\potts{local}).
\item What are generic strategies that can overcome the adversarial effect of
  predicted full annotations beyond the baseline, and how do they compare?
\begin{itemize}
\item Image-level regularisation using FC-CRF or Potts always helps
  (\potts{local}, \potts{global},\fccrf{global}, \potts{combined},
  \fccrf{combined}), up to a certain extent: improving the accuracy after
  training by one point requires a $5\times$ higher increase in the annotation
  quality (see linear trends in Fig.~\ref{fig:overview}).
\item Averaging local and global predicted annotations (\soft{combined}) leads
  to better annotations than regularizing the predictions of the baseline
  (\potts{global},\fccrf{global}). The combined predictions are less adversarial
  for training a segmentation network and it achieves state-of-the-art
  performances on this problem.
\end{itemize}
\end{itemize}

\clearpage

\section{Appendix}

In this supplementary, we provide detailed architectures and training
strategies, that we used to produce the results in the main paper.

\subsection{DeepLab: Architecture and Training}

Our strategy involves DeepLabV2 in two different tasks: (i) as a global
\gls{pam} and (ii) as a final segmentation \gls{dcnn} that is trained either on
weak, predicted or human full annotations.
For both tasks, we use the same architecture and training strategy.
As a global \gls{pam}, we disable both multi-scaling and \gls{fc-crf}
post-processing~\cite{Krahenbuhl2011}, unless mentioned otherwise.
However, as the final segmentation \gls{dcnn}, we assess the effect of both
multi-scaling and post-processing by testing all combinations, see Tab.~2 in the
main text.

The architecture of DeepLabV2 is based on ResNet-101, connected with an
\gls{aspp} module with four branches (\emph{atrous} rates
$r \in \Set{6, 12, 18, 24}$) and a bilinear up-sampling to match the input
resolution.
We initialise the model using the parameters of ResNet-101 pre-trained on
ImageNet~\cite{Russakovsky2015}.
We use random initialisation for all other parameters. In particular, we avoid
pre-training DeepLabV2 on any segmentation dataset such as
\acrshort{ms-coco}~\cite{Chen2016b}.
We modify a publicly available TensorFlow implementation of
DeepLabV2~\cite{Wang2018}.
Our training is as follows: 20k training iterations, a batch size of 10, a
momentum of 0.9, a weight decay of $0.0005$ and the following \emph{poly}
learning rate policy~\cite{Chen2016b}
$(1 - \mathrm{iter}/\mathrm{maxiter})^{\mathrm{power}}$, with a power of 0.9 and
a learning rate of $0.000625$.
We also enable data augmentation by randomly mirroring and scaling the input
data.
We calculate the loss after bilinear up-sampling, (input resolution
$321\times321$). We keep the same training policy for all annotation types
(weak, predicted or human full annotations).

\subsection{PSPNet}

The architecture of the PSPNets~\cite{Zhao2017} is based on either ResNet-50 or
ResNet-101.
We initialize the models using the parameters of the ResNets pre-trained on
ImageNet~\cite{Russakovsky2015}.
Again, we avoid pre-training on any other data set, especially on segmentation
data sets.
We use a publicly available TensorFlow implementation of
PSPNet~\cite{Li2019tensorflow}.

Our training is as follows: 30k training iterations, a batch size of 16, a
$L^2$--$\textrm{SP}$~\cite{Li2018} regularisation with parameters $\alpha=0.001$
and $\beta = 0.0001$, a momentum optimizer with a momentum of 0.9 and the
following \emph{poly} learning rate policy~\cite{Chen2016b,Zhao2017}
$(1 - \mathrm{iter}/\mathrm{maxiter})^{\mathrm{power}}$, with a power of 0.9 and
a learning rate of 0.01.
We use random rotation and random scaling for data augmentation with an image
resolution of $480 \times 480$.
We keep the same training policy for all annotation types (weak, predicted or
human full annotations).

\subsection{Potts}
We select the regularisation parameter $\regp$ by a grid search on a small
subset of the training set.
We use the parameters that achieve the highest \gls{miou} on the predicted
annotations: $\regp = 10$ when using the local or combined \glspl{pam} and
$\regp = 50$ when only the global \gls{pam} is used.
For the edge stop function, we use the parameter $\eta = 0.01$.

\subsection{\acrlong{fc-crf}}
We selected the parameters of the \gls{fc-crf} by a grid search on the predicted
full annotations, obtained from the combined model applied to a small subset of
the training set.
We use the same parameters for all experiments: $w_1 = 3$, $\sigma_\alpha=30$,
$\sigma_\beta=5$, $w_2=5$ and $\sigma_\gamma=2$.

%%% Local Variables:
%%% mode: latex
%%% TeX-master: "egpaper"
%%% End:

{\small
\bibliographystyle{ieee}
\bibliography{egbib}
}

\end{document}